\documentclass{esannV2}
\usepackage{graphicx}
\usepackage[utf8]{inputenc}
\usepackage{amssymb,amsmath,array}
\usepackage{hyperref}
\usepackage{booktabs}
\usepackage{tikz}
\usepackage[english]{babel}
\usetikzlibrary{positioning,shapes.geometric,arrows.meta,calc}
\usepackage{xcolor}
\usepackage{subcaption} 

\begin{document}
\title{See Without Decoding: Motion-Vector-Based Tracking in Compressed Video}

\author{Duch\'e Axel$^{1,2}$, Chatelain Cl\'ement $^2$ and  Gasso Gilles$^2$
%
\thanks{This research is supported by ANRT and ACTEMIUM Paris Transport. We thank ACTEMIUM Paris Transport for the funding and CRIANN for the computation facilities}
%
\vspace{.3cm}\\
%
1- Actemium Paris Transport, 
Nanterre - France\\
%
2- INSA Rouen, LITIS UR 4108, Saint-Étienne-du-Rouvray - France
}

\maketitle

\begin{abstract}
We propose a lightweight \textbf{compressed-domain tracking model} that operates directly on video streams, 
without requiring full RGB video decoding. Using motion vectors and transform coefficients from compressed data, our deep model propagates object bounding boxes across frames, 
achieving a computational speed-up of order up to $3.7\times$ with only a slight 4\% mAP@0.5 drop vs RGB baseline on MOTS15/17/20 datasets. These results highlight codec-domain motion modeling efficiency for real-time analytics in large monitoring systems.
\end{abstract}
\vspace{-2mm}
\section{Introduction}
\vspace{-2mm}
\label{sec:intro}

Modern cities operate extensive camera networks across public areas and sensitive zones. These systems must deliver reliable, continuous analytics---including motion detection, intrusion monitoring, and behavior analysis---under strict constraints on computation, storage, and energy consumption across thousands of concurrent video streams. Conventional image-domain preprocessing~\cite{PIDS_Survey_2022} (e.g., background subtraction or frame differencing) offers low-cost activity filtering but remains fragile under illumination or perturbations. Deep learning--based vision models have substantially improved robustness and precision while maintaining a good balance between speed and accuracy, as seen in recent architectures such as RT-DETR~\cite{detrs_beat_yolos_2024} and the latest YOLOs~\cite{khanam2024yolov11overviewkeyarchitectural}. However, most vision pipelines and methods still rely on fully decoded RGB frames, which are computationally heavy to process. Performing real-time inference on high-resolution RGB data requires powerful GPUs or specialized hardware, making large-scale deployment difficult. The reliance on extensive RGB processing poses a significant scalability challenge for camera networks. This paper adress this limitation by starting from a simple hypothesis: Compressed video streams already carry most of the spatial–temporal information required for tracking. Based on this, this work proposes a lightweight hybrid architecture that performs a single detection on an initial decoded RGB frame, followed by tracking and refinement of bounding boxes directly from codec-domain features. Hence, the method avoids redundant pixel-level computation while maintaining competitive accuracy and supporting scalable, energy-efficient analytics across large camera infrastructures. Hereafter, we first explain the compressed video representation that enables our approach and review prior work on compressed-domain video understanding and how these methods balance efficiency and accuracy under similar constraints.

\vspace{-1mm}
\section{Compressed Video and Related Work}
\label{sec:sota}
\vspace{-2mm}
\subsection{Compressed Video Representation}
\label{subsec:compressed_representation}

Standard video codecs (e.g., MPEG-4, H.264) organize streams into \emph{Groups of Pictures} (GOPs) $\mathcal{G}^g=\{f_0^g, f_1^g, \ldots, f_N^g\}$, where $f_0^g$ is a fully encoded RGB image, termed I-frame, and temporal frames $f_n^g$ ($n\geq1$), also known as P-frames, are encoded via block motion estimation.
Each P-frame contains two components:
$
f_n^g = \{MV_n^g, \Delta (Y,Cb,Cr)_n^g\},
$
Motion vectors $MV_n^g$ that capture block-wise displacement by seeking matching blocks within a search zone, while residuals $\Delta (Y,Cb,Cr)_n^g$ encode appearance differences via pixel subtraction.
Since MPEG-4 applies Discrete Cosine Transform (DCT) to compress residuals, one has access to directly work with 
DCT coefficients that compactly represent texture, edges, and appearance changes in the frequency domain.

These "codec features" are already computed during encoding, making them extractable from the bitstream at minimal cost—a key advantage for large-scale video analytics.  
Motion vectors are dramatically more compact than RGB images because they encode the displacement of a 16×16 pixel block using only two floating-point values. DCT residuals, on the other hand, remain considerably larger, yet still far lighter than full RGB data, providing efficient motion and appearance cues at negligible extraction cost.

\subsection{Related Work and Positioning}

In this work, we exploit codec features to design a fast tracking models with minimal decoding. Previous studies have explored different levels of
reconstruction before prediction. 

\emph{Extensive decoding:}
Conventional approaches fully reconstruct RGB frames and increase the inputs with compressed features. 
Methods such as MV-YOLO~\cite{mv_yolo_motion_vector_2018}, originally designed for city surveillance, and later RGB--motion fusion variants like MV-Soccer~\cite{majeed2024mvsoccer} and ReST~\cite{majeed2025rest}, for sports tracking, all exploit codec motion vectors alongside RGB cues to enhance temporal consistency.  
Although this design effectively leverages motion information, it still depends on full-resolution RGB decoding and deep feature extractors, resulting in architectures that are computationally demanding.

\emph{Partial decoding:} To reduce overhead, a few methods leverage motion and residual information computed by the codec while still decoding parts of the frame.  Frame-group aggregation methods~\cite{wang2019fastobjectdetectioncompressed,Wang_2021} process an entire GOP in a single inference using the initial decoded RGB I-frame and decompressed
residuals and motion vectors to predict the objects in all the frames of the GOP.Despite the use of smaller data, this architecture is still computationally intensive and offers only limited improvements in terms of runtime efficiency.
Other hybrid designs~\cite{Fast_Object_Detection_in_High-Resolution_Videos_2023} decode small regions or key patches while combining them with motion cues from the bitstream.  
However, because video data are sequentially encoded, even partial decoding typically requires decoding earlier related patches, which limits the achievable efficiency.

\emph{No-decoding methods:} Recent works such as \cite{dct_luminance_2021} and \cite{Gueguen2018} operate directly on compressed-domain, bypassing the need for  RGB reconstruction. Video-domain approaches are reliant on codec features alone. One example is the naïve Mean-MV baseline in ~\cite{Fast_Object_Detection_in_High-Resolution_Videos_2023}, where boxes obtained from an RGB image are tracked using the average of the motion vectors in each box. There's also the MV models~\cite{Deg23} and DCT-based methods~\cite{elkhoury2021deep}.  Although highly efficient, they struggle with static or slow-moving objects due to a lack of motion information. Nevertheless, no-decoding strategies remain attractive for scalable and efficient video analytics.

Building on the latter idea, we propose a hybrid lightweight tracking scheme that bridges partial decoding and no-decoding, where only the I-frame is decoded, while therest of the GOP is being handled entirely in the compressed domain. This eliminates the need for repeated reconstruction while ensuring that the cues needed for accurate, scalable video analytics are maintained.

\vspace{-2mm}
\section{See Without Decoding using BAFE}
\label{sec:architecture}
\begin{figure}[h!]
\centering
\includegraphics[width=0.90\linewidth]{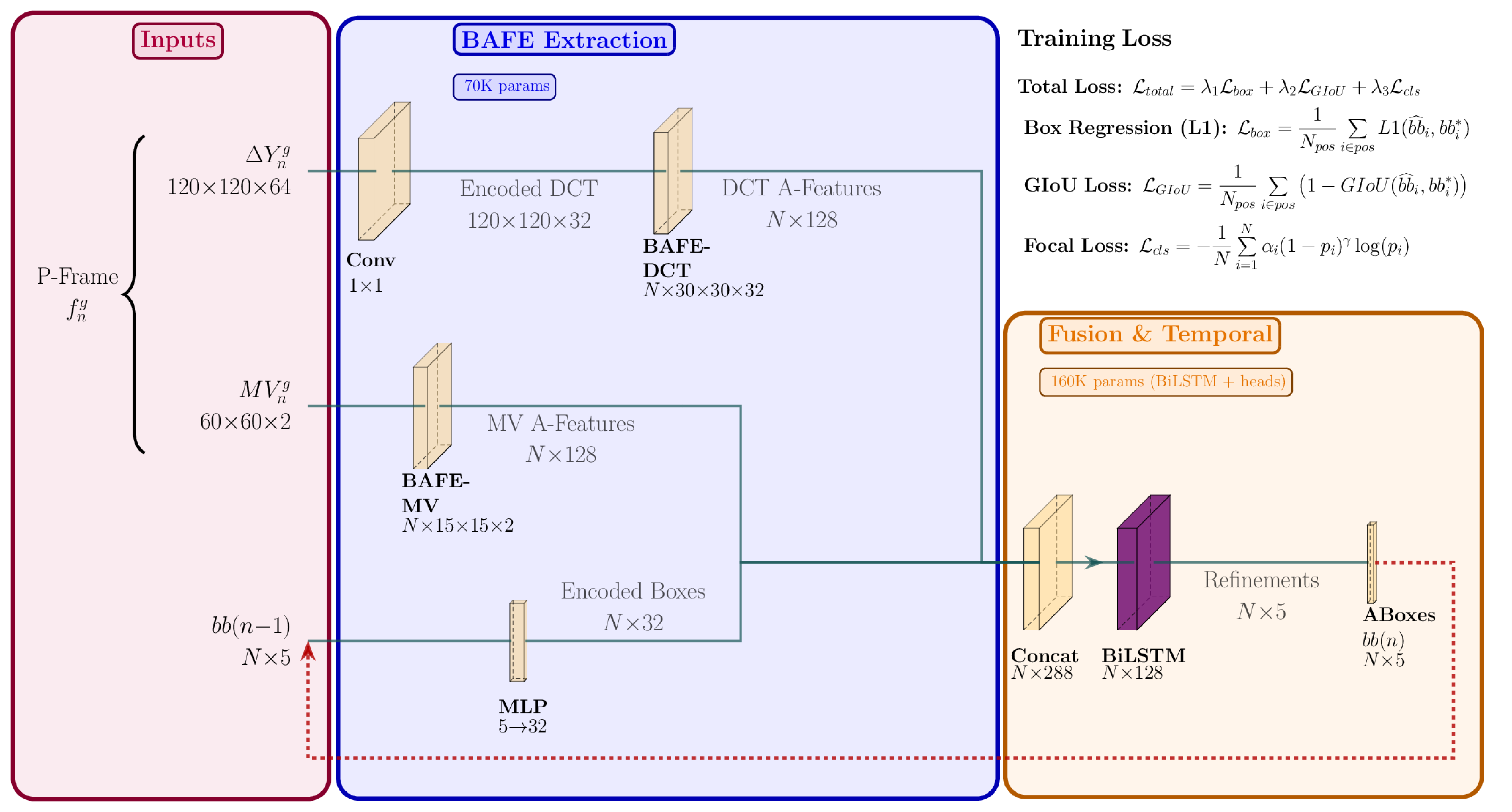}
\caption{BAFE architecture: box-aligned MV/DCT features are fused with box encodings and then fed to BiLSTM for temporal modeling.}
\label{fig:roi_architecture}
\end{figure}

\vspace{-2mm}
Our method involves a single RGB detection step for the I frame and a fully compressed-domain tracking system for the P frames, which aims to propagate the initial detections alongside the subsequent motion vectors. This results in a lightweight hybrid architecture.  As shown in Figure~\ref{fig:roi_architecture}, once the initial boxes are obtained, each P-frame $f_n^g$ provides its motion vectors $MV_n^g$, DCT residuals $\Delta Y_n^g$, and the previous boxes $bb(n\!-\!1)$ at the \emph{Inputs} stage (left). To replace naïve motion-only propagation such as Mean-MV, we introduce a \emph{Box-Aligned Feature Extraction (BAFE)} module, which gathers motion and appearance cues from both the box and its surrounding neighborhood. In the \emph{BAFE Extraction} block (center), motion vectors and DCT residuals are transformed into compact MV and DCT A-Features, while previous boxes are encoded through a lightweight box embedding. These codec features are then fed into the \emph{Fusion \& Temporal} module (right), where a BiLSTM predicts refinement terms that the \emph{ABoxes} component transforms into boxes $bb(n)$ by updating previous boxes $bb(n-1)$.

\textbf{Addressing SOTA Limitations.}
Our design directly tackles limitations previously observed.
(1) \emph{vs.\ partial- and full-decoding methods}~\cite{wang2019fastobjectdetectioncompressed,Wang_2021,mv_yolo_motion_vector_2018,majeed2024mvsoccer}: our approach avoids any form of repeated or selective RGB reconstruction, eliminating both the heavy computational cost associated with full decoding and the synchronization overhead inherent to partial decoding;
(2) \emph{vs.\ no-decoding methods}~\cite{elkhoury2021deep,dct_luminance_2021}: by performing a single RGB-based detection on the I-frame $f_0^g$, we obtain high-quality initial boxes that allow us to track static and slow-moving objects, which motion-only approaches typically fail to capture.
The strong I-frame priors convert a difficult detection problem into a lightweight propagation task for which codec features are sufficiently informative. To reliably exploit these compressed-domain cues, we further introduce the BAFE module, which provides a learnable alternative to naïve MV-based propagation by extracting richer motion and appearance information from both the interior of each bounding box and its immediate neighborhood. To reduce computational cost, we operate only on the luminance (Y) for residuals, cutting data volume by 3$\times$ while preserving spatial-temporal structure for tracking.

\textbf{BAFE (Box-Aligned Feature Extraction).}
The BAFE module extracts features \emph{spatially aligned} with each bounding box, ensuring that only object-relevant regions are processed. For every box $b_n^i$ in frame $f_n^g$, motion vectors $MV_n^g$ and DCT residuals $\Delta Y_n^g$ are sampled on a fixed-size grid. Because block motion estimation in \emph{MPEG-4 Part 2} operate on fixed block size. BAFE selects a compact grid covering both the box interior and its immediate neighborhood, capturing the most informative motion and appearance cues. The chosen grid sizes (Figure~\ref{fig:roi_architecture}) balance detail and efficiency, producing lightweight MV and DCT A-Features while retaining essential spatiotemporal structure. By focusing computation on box-aligned regions instead of full-frame tensors, BAFE provides an expressive yet efficient representation well suited for real-time tracking.

\textbf{BiLSTM Temporal Propagation.} To model temporal dependencies across P-frames, we employ a bidirectional LSTM that takes as input the concatenation of: (1) box-aligned motion vector features from $MV_n^g$, (2) box-aligned DCT coefficient features from $\Delta Y_n^g$, and (3) encoded bounding box geometry (position/size of $b_{n-1}^i$). The BiLSTM hidden state captures motion patterns over time, enabling the model to predict refined bounding box updates $\Delta b_n^i = (\Delta x, \Delta y, \Delta w, \Delta h)$ relative to the previous frame. This temporal modeling is crucial for handling occlusions, camera motion, and non-linear trajectories that simple motion vector averaging cannot capture.

\textbf{ABoxes and Training Loss.}
The ABoxes module applies the predicted refinements to the previous frame's bounding boxes by summing the predictions with the previous coordinates. Training relies on a weighted combination of three standard detection objectives: an L1 loss ($\mathcal{L}_{\text{box}}$) for bounding box regression, a GIoU loss ($\mathcal{L}_{\text{GIoU}}$) to improve geometric alignment, and a focal loss ($\mathcal{L}_{\text{cls}}$) to manage potential disappearances. The total loss is computed as $\mathcal{L}_{\text{total}} = \lambda_1 \mathcal{L}_{\text{box}} + \lambda_2 \mathcal{L}_{\text{GIoU}} + \lambda_3 \mathcal{L}_{\text{cls}}$ with weights $\lambda_1 = 5$, $\lambda_2 = 2$, and $\lambda_3 = 2$. The box regression and GIoU losses on boxes by comparing predicted boxes $\widehat{bb}_i$ with ground truth $bb_i^*$. The focal loss is computed over all $N = 200$ proposals using class balance weights $\alpha_i$, predicted probabilities $p_i$, and focusing parameter $\gamma = 2$ to down-weight easy examples and focus training on hard negatives.
\vspace{-2mm}
\section{Experiments and results}
\vspace{-2mm}
We train and evaluate on MOT15 \cite{MOTS15}/17 \cite{MOTS17}/20\cite{MOTS20}  with baselines: RGB-based detection per frame and \textbf{Mean-MV} also used in \cite{Fast_Object_Detection_in_High-Resolution_Videos_2023}. The MOT datasets focus on pedestrian tracking from surveillance camera recordings and on-board camera but we choose to only work on static camera as our main application is surveillance. The videos used for training and evaluation are static camera videos from the same train and test sets defined in the 3 MOTS datasets. All trainable models have been trained on 200 GOPs of size 6 for 100 epochs.

We report mAP@0.5 since the tracker runs without IDs and aims to minimize detector usage. As shown in Table~\ref{tab:results_combined}, our best BAFE model (MV+DCT) reaches \textbf{0.8962 mAP} on MOT17, improving over Mean MV by \textbf{+2.86\%} and staying within \textbf{2.25\%} of the RGB baseline. On MOT15 and MOT20, it achieves gains of \textbf{over 20\%} by better capturing fast pedestrian motion.

\begin{table*}[h!]
\centering
\scriptsize
\setlength{\tabcolsep}{6pt}
\renewcommand{\arraystretch}{0.9}

\caption{BAFE model performance and deployment scalability on static cameras (GOP-6, mAP@0.5, 30 FPS).}
\label{tab:results_combined}

\begin{tabular}{lccc || cc}
\toprule
\textbf{Model} & \textbf{MOT17} & \textbf{MOT15} & \textbf{MOT20} 
& \textbf{concurrent streams} & \textbf{Params} \\
\midrule

Baseline RT-DETRv2 & 0.9187 & 0.8234 & 0.8153 & 13 & 31.450M \\

Mean MV & 0.8676 & 0.5879 & 0.6590 & \textbf{85} & \textbf{0} \\

ours BAFE (MV) & 0.8756 & 0.7851 & 0.7990 & 48 & 160K \\

ours BAFE (DCT) & 0.8543 & 0.7564 & 0.7843 & 45 & 202K \\

ours BAFE (MV+DCT) & \textbf{0.8962} & \textbf{0.7958} & \textbf{0.8020} & 42 & 230K \\

\bottomrule
\end{tabular}

\vspace{-0.5em}
\end{table*}

\emph{Deployment Scalability.}
Table~\ref{tab:results_combined} shows deployment capacity on 16GB GPU for GOP-6 (1 I-frame + 5 P-frames). Despite its compact 160K parameters, our MV-only variant is able to process \textbf{48 concurrent streams} vs. \textbf{13 streams} for RT-DETR (32M params), achieving \textbf{3.7$\times$ throughput} with 200$\times$ fewer parameters. The naive Mean MV baseline handles 85 streams but sacrifices 20\% mAP compared to our learned model.

\emph{Discussion.}
Our BAFE-based models consistently outperform the naïve Mean-MV baseline, confirming the benefit of learned box-aligned compressed features combined with temporal modeling. The MV-only variant achieves strong accuracy with high throughput, illustrating the efficiency of motion-vector–based tracking. We fix the GOP size to 6 to preserve accuracy while keeping the propagation window manageable; larger GOPs would require a heavier temporal model and would further expose the limitation that new object appearances within a GOP are not handled. Our approach also assumes static cameras and relies on codec motion consistency, which can reduce performance under significant camera motion or highly irregular residual patterns, though these cases remain limited in our target surveillance scenarios.

\vspace{-1mm}
\section{Conclusion and Future Work}
\vspace{-1mm}
We presented a hybrid compressed-domain tracking framework that performs a single RGB-based initialization before relying entirely on codec-domain features. Through the BAFE module and a lightweight temporal head, the method achieves strong accuracy while maintaining high scalability, demonstrating the advantages of exploiting compressed-domain information directly. Future work includes extending BAFE to multi-scale box-aligned grids to improve robustness to object scale variations. We also plan to improve our BiLSTM to increase the size of the GOP that we can tackle without performance loss.


\begin{footnotesize}





\end{footnotesize}


\end{document}